\documentclass[sigconf]{acmart}
\renewcommand\footnotetextcopyrightpermission[1]{} 
\AtBeginDocument{%
  }

\copyrightyear{}
\acmYear{}
\acmDOI{}
\acmConference[]{}
\acmISBN{}

\usepackage{multirow}
\usepackage{graphicx} 
\usepackage{float}
\usepackage{wrapfig}

\begin{document}
\title{	Towards Generalizable Deepfake Detection with Spatial-Frequency Collaborative Learning and Hierarchical Cross-Modal Fusion}

\author{Mengyu Qiao}
\authornote{Corresponding author.}
\affiliation{%
  \institution{ North China University of Technology}
  \city{Beijing}
  \country{China}
}
\email{myuqiao@ncut.edu.cn}

\author{Runze Tian}
\affiliation{%
  \institution{North China University of Technology}
  \city{Beijing}
  \country{China}}
\email{trz@mail.ncut.edu.cn}

\author{Yang Wang}
\affiliation{%
  \institution{Ultramain Systems, Inc.}
  \city{Albuquerque}
  \state{NM}
  \country{USA}}
  \email{ywang@ultramain.com}

\renewcommand{\shortauthors}{}

\begin{abstract}
 The rapid evolution of deep generative models poses a critical challenge to deepfake detection, as detectors trained on forgery-specific artifacts often suffer significant performance degradation when encountering unseen forgeries. While existing methods predominantly rely on spatial domain analysis, frequency domain operations are primarily limited to feature-level augmentation, leaving frequency-native artifacts and spatial-frequency interactions insufficiently exploited. To address this limitation, we propose a novel detection framework that integrates multi-scale spatial-frequency analysis for universal deepfake detection. Our framework comprises three key components: (1) a local spectral feature extraction pipeline that combines block-wise discrete cosine transform with cascaded multi-scale convolutions to capture subtle spectral artifacts; (2) a global spectral feature extraction pipeline utilizing scale-invariant differential accumulation to identify holistic forgery distribution patterns; and (3) a multi-stage cross-modal fusion mechanism that incorporates shallow-layer attention enhancement and deep-layer dynamic modulation to model spatial-frequency interactions. Extensive evaluations on widely adopted benchmarks demonstrate that our method outperforms state-of-the-art deepfake detection methods in both accuracy and generalizability.
\end{abstract}



\renewcommand\footnotetextcopyrightpermission[1]{}
\settopmatter{printacmref=false} 

\maketitle

\section{Introduction}
The rapid advancement of deep-generative models \cite{arjovsky2017bottou}\cite{karras2017progressive}\cite{karras2019style} has catalyzed the evolution of facial manipulation techniques, enabling the creation of highly realistic synthetic media (commonly termed "deepfakes"). Although these technologies \cite{stargan}\cite{attgan}\cite{faceshifter}\cite{face2face} enable potential applications in the entertainment and creative industries, their misuse poses significant social risks, including disinformation propagation, identity theft, and political fraud. Consequently, detecting deepfake content has emerged as a critical challenge in multimedia security and computer vision.

Most existing deepfake detection methods \cite{x-ray}\cite{efficientnet}\cite{xception}\cite{mesonet} formulate the task as binary classification, where backbone networks are typically used to extract discriminative features. Analysis and detection \cite{mad}\cite{wang2023noise}\cite{nirkin}\cite{yin2023dynamic}\cite{zhang2022deepfake} predominantly focus on artifacts in the spatial domain, such as facial inconsistencies, unnatural textures, blending boundaries, etc. However, as sophisticated deepfake techniques increasingly refine spatial details, the distinctions between authentic and forged faces at the pixel level have become remarkably subtle and undetectable. This limitation reveals the inherent vulnerability of spatial domain approaches, particularly when confronting high-quality forgeries or post-processing perturbations.

\begin{figure}[htbp]
    \centering 
    \vspace*{10pt}
    \includegraphics[width=0.4\textwidth]{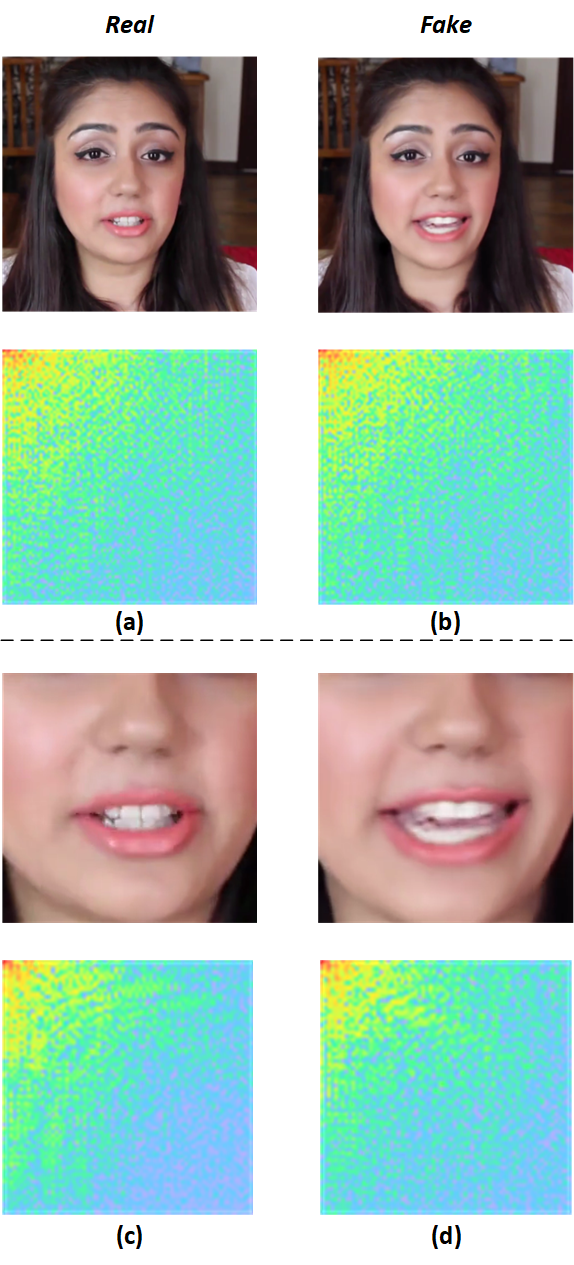} 
    \caption{(a) (b) represent heatmaps of Full-Image DCT transform and (c) (d) represent heatmaps of Local Regional DCT Transform between real and fake images.} 
    \label{fig:hotmap} 
\end{figure}

In contrast, frequency domain analysis offers a complementary perspective for uncovering manipulation traces. Forged faces often exhibit distinctive spectral patterns, such as inconsistencies in high-frequency noise or anomalies introduced by upsampling operations in generative models. Despite this potential, existing frequency-aware methods \cite{sstnet}\cite{agarwal2021md}\cite{tan2024frequency}\cite{wavelet}\cite{luo2021generalizing} typically augment spatial anomalies through intermediate filtering operations in spectral representations and finalize feature extraction in spatial domain, neglecting the intrinsic abnormal patterns manifested across frequency sub-bands, especially in the local forged regions, as shown in Figure \ref{fig:hotmap}. Although several works \cite{f3net}\cite{blockdct} incorporate block-wise frequency transformations, they fail to deeply investigate the interdependencies among frequency features. Furthermore, most methods inadequately address the dependence between image semantics and forgery artifacts, resulting in limited generalization across datasets and manipulation types.

To address these challenges, we propose a Spatial-Frequency Collaborative Learning (\textbf{SFCL}) framework that simultaneously extracts features from both spatial and frequency domains while enabling multi-stage feature interaction through a Hierarchical
Cross-Modal Fusion (\textbf{HCMF}) mechanism. For frequency domain processing, we depart from conventional global frequency transformation and instead adopt a localized analysis paradigm. Specifically, we employ block-wise Discrete Cosine Transform (DCT) to precisely capture frequency domain artifacts while retaining macro-level spatial semantics. To model the distribution pattern of intra-block DCT coefficients and the transitional patterns of inter-block DCT sub-bands, we construct a local branch in the frequency pipeline that extracts multi-scale frequency features through cross-band and cross-block convolution. Additionally, we introduce a global branch in the frequency pipeline, to compute statistical frequency features from horizontal and vertical derivatives of DCT sub-bands across blocks of a face patch at the original resolution. Finally, we incorporate a frequency-guided attention mechanism to enhance spatial features in the shallow layers of the network while enabling effective cross-modal fusion in the deep layers.

Extensive experiments on benchmark datasets (FaceForensics++, Celeb-DF, and DFDC) validate the superior performance of our method in both intra-dataset evaluation and cross-dataset generalization. In summary, the contributions of this study are threefold as below:

\begin{itemize}
    \item We propose a novel block-wise discrete cosine transform analysis approach with inter/intra-block multi-scale frequency-convolutional network. Through progressive feature abstraction in the DCT transform domain, this approach effectively captures both intra-block spectral characteristics and inter-block transitional patterns, thereby enhancing local representation learning and precise artifact modeling in the frequency domain.
\end{itemize}

\begin{itemize}
    \item  We propose a global spectral feature extraction and synthesis approach based on scale-invariant differential analysis. This approach models holistic forgery patterns across regions of interest while minimizing multi-scale discrepancies and reducing interference from post-processing operations.
\end{itemize}

\begin{itemize}
    \item We develop a hierarchical cross-modal fusion mechanism that integrates shallow-layer attention enhancement and deep-layer dynamic modulation. This design effectively models spatial-frequency interactions, addresses modality-specific feature heterogeneity in detection tasks, and maintains robustness against sophisticated forgery manipulations.
\end{itemize}

\section{Related Works}
Early detection methods focused on the unnatural biological signals in forgeries, such as abnormal mouth movements during speech \cite{agarwal2020detecting},head poses \cite{yang2019exposing}, irregular blink frequency and anomalies in remote visual photoplethysmography (rPPG) \cite{ciftci2020fakecatcher}\cite{hernandez2020deepfakeson}\cite{qi2020deeprhythm}\cite{fernandes2019predicting}.

With the advancement of deep learning, various models \cite{xception}\cite{efficientnet}\cite{mesonet} have been widely adopted as backbone networks to detect deepfake in the spatial domain. To further improve performance, many methods \cite{mad}\cite{dynamicgraph}\cite{ba2024exposing}\cite{nguyen2024laa} incorporate specialized functional modules to enhance the learning of localized forgery patterns. For instance, M2TR \cite{m2tr} and HFI-Net \cite{hfi-net} leverage the complementary strengths of Vision Transformers (ViTs) \cite{vit} and convolutional neural networks (CNNs) to boost detection accuracy. MAD \cite{mad} introduces a multi-attentional architecture equipped with texture enhancement blocks and a regional independence loss, allowing the network to attend to non-overlapping regions and capture fine-grained local forgery cues. Nirkin et al. \cite{nirkin} propose a method that exploits discrepancies between facial regions and their surrounding context, based on the observation that most manipulation techniques modify only the face while preserving peripheral features such as hair, ears, and neck. Additionally,  some approaches detect manipulations by integrating noise features \cite{zhang2024face}\cite{wang2023noise}\cite{luo2021generalizing}\cite{zhou2017two} with RGB features.

Recent studies \cite{durall}\cite{f3net}\cite{wavelet}\cite{twobranch}\cite{spsl}\cite{exploiting}\cite{liu2024attention} reveal the promising effects of discriminative patterns in the frequency domain for deepfake detection. Qian et al. \cite{f3net} proposed F3-Net, separating frequency components via DCT and fusing local frequency statistics, achieving robustness on compressed videos. Durall et al. \cite{durall} analyzed high-frequency spectral anomalies using Discrete Fourier Transform (DFT), providing theoretical insights into frequency-domain forgery patterns. Several approaches leverage the complementarity between the frequency domain and RGB features. Masi et al.\cite{twobranch} designed a dual-stream network combining Laplacian-of-Gaussian frequency enhancements with spatial features. Liu et al. \cite{spsl} jointly modeled upsampling artifacts using phase spectra and spatial Moreover, spatial-frequency relation \cite{dynamicgraph}\cite{hfi-net}\cite{exploiting}\cite{fang2025deepfake} was studied to leverage the representational capacity of multi-modal learning. Although these methods utilize frequency analysis, most of them simply rely on filter-based approaches to enhance spatial domain detection, lack in-depth exploration of diverse frequency characteristics at local and global scales, and neglect the interactions of the spatial and frequency domains at different stages.

\section{Methods}
\subsection{Overview}
\begin{figure*}[]
    \centering 
    \vspace*{10pt}
    \includegraphics[width=2\columnwidth]{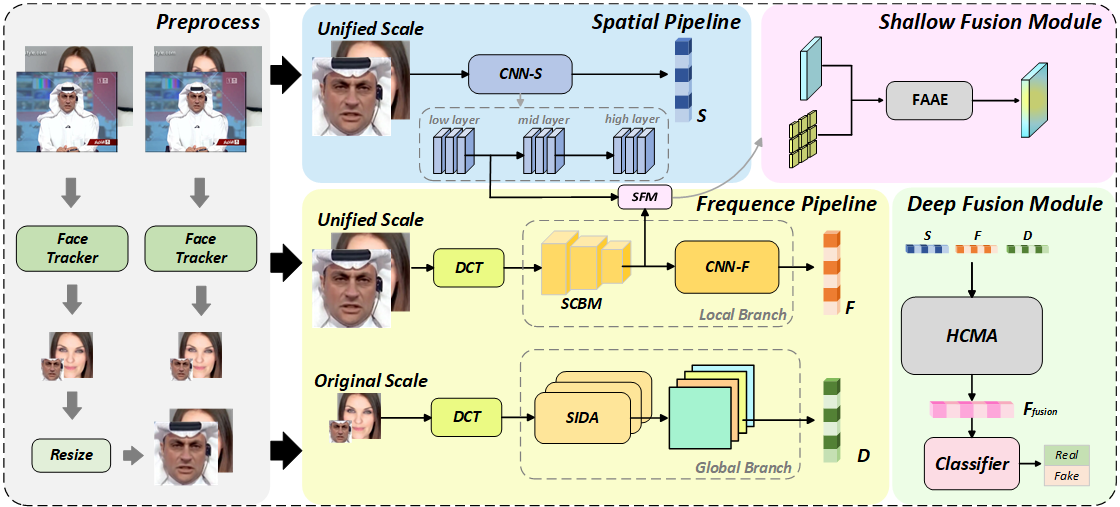} 
    \caption{The architecture of SFCL-HCMF.} 
    \label{fig:framework} 
\end{figure*}

This section presents the design motivations and architectural overview of our proposed framework. Modern deepfake techniques progressively synthesize content from local to global scales, often resulting in inter-block pattern inconsistencies, particularly around semantic boundaries, such as facial regions. To address diverse patterns of deepfake artifacts, we employ a Spatial-Frequency Collaborative Learning paradigm to achieve comprehensive representation and generalizable detection. An overview of our proposed framework is illustrated in Figure \ref{fig:framework}.

In the frequency domain, deepfake manipulations tend to introduce additive noise or unnatural spectral distortion during the synthesis of fine-grained details. We hypothesize that applying block-wise DCT transform enables direct analysis of high-, mid-, and low-frequency artifacts within each local patch. By partitioning images into blocks, our method focuses on localized spectral characteristics, thereby mitigating the detail loss inherent in global transformations while enhancing sensitivity to subtle anomalies.

We identify the limitations of popular backbone network architectures in frequency domain analysis. Specifically, standard 2D-CNN backbones are insufficient for capturing deep interdependencies among DCT coefficient distributions within image blocks and lack the representational capacity for effective inter-block frequency analysis. Moreover, many existing methods rely on upsampling or downsampling operations (e.g., sinc or bilinear interpolation) to align input dimensions, which often degrades frequency-domain artifacts and diminishes discriminative performance.

Motivated by these observations, we propose a Local-Global Frequency Framework within the frequency analysis pipeline. This framework integrates two complementary branches:
Local Branch: We design an inter/intra-block multi-scale frequency-convolutional network to hierarchically model intra-block spectral coefficient correlations and inter-block transitional patterns, enabling localized frequency feature extraction.
Global Branch: We introduce Scale-Invariant Differential Analysis (\textbf{SIDA}), applied to cropped regions of interest in original resolution. This approach avoids high-frequency degradation caused by resampling and preserves scale-invariant spectral fingerprints critical for global pattern analysis.

Moreover, we also incorporate a spatial domain analysis pipeline by utilizing an EfficientNet backbone network which is pre-trained on image classification tasks and fine-tuned on deepfake detection tasks to capture local spatial artifacts, such as textures, edges, and structural inconsistencies.

To bridge the frequency and spatial domains, we further propose a Hierarchical Cross-Modal Fusion Framework. In the shallow layers, we incorporate a Frequency-Aware Attention Enhancement Module (\textbf{FAAE}), which generates block-wise, frequency-localized attention maps from spectral features. This enhances spatial representations by highlighting localized forgery cues. In the deeper layers, we implement a Hybrid Cross-Modal Attention Fusion Module, which combines hybrid attention mechanisms for spatial–frequency coherence with differential-enhanced gating for anomaly amplification. Together, these modules enable the effective integration of multi-modal representations through hierarchical interactions.

\subsection{Local-Global Frequency Framework}
\begin{figure}[htbp]
    \centering 
    \vspace*{10pt}
    \includegraphics[width=0.5\textwidth]{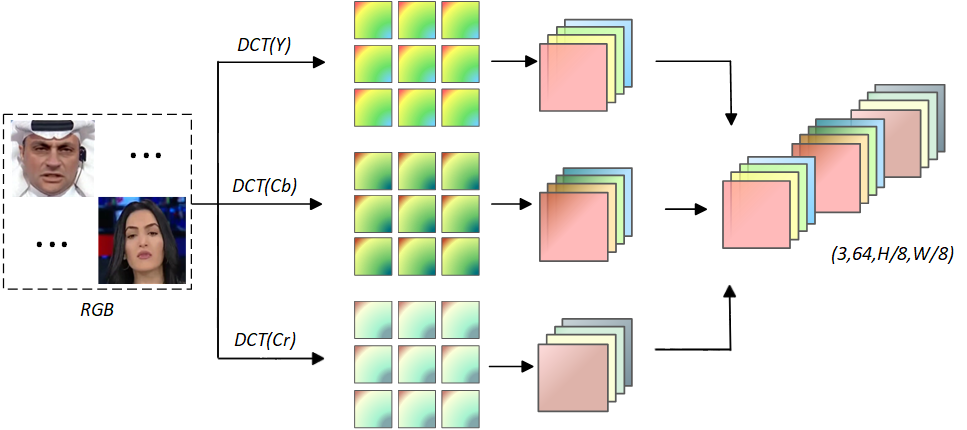} 
    \caption{The block-wise DCT transform.} 
    \label{fig:dct} 
\end{figure}

Formally, let the input image be denoted as $X\in \mathbb{R} ^{C\times H\times W}$, where H and W represent the height and width, respectively, and $C$ denotes the number of channels. Following the approach in \cite{dctnet}, we first convert the RGB image to the YCbCr color space. Subsequently, block-wise DCT transforms with a block size of 8×8 are applied across all three channels, consistent with standard image compression protocols such as JPEG. Each block of DCT coefficients is then flattened into a one-dimensional vector using zigzag scanning, which orders the coefficients from low to high frequency. This process aligns coefficients corresponding to the same frequency components across channels and systematically organizes all blocks based on their frequency content. The resulting representation is a restructured 4D tensor  $\tilde{X}  \in \mathbb{R} ^{C\times 64\times H/8\times W/8} $, which explicitly preserves spatial–frequency relationships crucial for subsequent analysis.As shown in Figure \ref{fig:dct}.

\subsubsection{Global Branch} 

Conventional preprocessing pipelines typically resize input images to fixed dimensions to facilitate batch processing. However, this operation inevitably alters the spectral energy distribution of forged regions and compromises the integrity of high-frequency information. To address this limitation, the global branch introduces the SIDA module that preserves block-wise DCT transformations at the original resolution while maintaining shape consistency of spectral features across images with varying spatial dimensions. Specifically, we first localize facial regions using a face tracking algorithm to obtain bounding box coordinates, within which block-wise DCT transformations are exclusively performed to retain region-specific spectral information.

For inter-block scale analysis, localized manipulations often cause abrupt variations in frequency distributions between adjacent row and column blocks. To capture transitional inconsistencies, we apply horizontal and vertical differential operations to eliminate semantic information while amplifying manipulation-specific anomalies. We compute inter-block differential operations on the frequency-transformed features $\tilde{X}$, deriving inter-block row differential features $\tilde{X}_{row}   \in \mathbb{R} ^{C\times 64\times (H/8)\times ((W/8)-1)}$ and inter-block column differential features 
$\tilde{X}_{col}   \in \mathbb{R} ^{C\times 64\times ((H/8)-1)\times (W/8)} $, which can be represented by equation 1 and 2.
\begin{equation}
   \tilde{X}_{row} = \tilde{X} _{:,:,m+1,:} - \tilde{X} _{:,:,m,:},  m \in[1,(H/8)-1]
\end{equation}
\begin{equation}
    \tilde{X}_{col} = \tilde{X} _{:,:,:,n+1} - \tilde{X} _{:,:,:,n} , n \in[1,(W/8)-1]
\end{equation}

At the intra-block scale, forged regions often display anomalous spectral energy distributions. Intra-block coefficient differentials quantify local correlations between adjacent frequency components. We then perform intra-block differential operations on $\tilde{X}$ to obtain intra-block differential features $\tilde{X} _{intra} \in \mathbb{R} ^{C\times 63\times (H/8)\times (W/8)}$ as represented in equation 3, which then undergoes zero-padding along the intra-block dimension to maintain dimensional consistency.
\begin{equation}
    \tilde{X}_{intra} = \tilde{X} _{:,l+1,:,:} - \tilde{X} _{:,l,:,:}, l \in[1,63]
\end{equation}

Subsequently, four statistical metrics—mean, std, skewness, and kurtosis, are computed on the absolute values of the three derivative feature maps, as shown in equation 4-7:
\begin{equation}
    {Mean}_{k} = \frac{1}{H\times W} \sum_{i=1}^{H} \sum_{j=1}^{W}  \left | {\tilde{X} }_{k}^{:,:,i,j} \right | ,(k=row,col,intra)
\end{equation}
\begin{equation}
    {Std}_{k} = \sqrt{\frac{1}{H\times W} \sum_{i=1}^{H} \sum_{j=1}^{W}  (\left | {\tilde{X} }_{k}^{:,:,i,j} -Mean_{k} \right |)^{2} } 
\end{equation}
\begin{equation}
    {Skew}_{k} = \frac{\frac{1}{H\times W} \sum_{i=1}^{H} \sum_{j=1}^{W}  (\left | {\tilde{X} }_{k}^{:,:,i,j} -Mean_{k} \right |)^3}{{{^{Std_{k}3} } }  }  
\end{equation}
\begin{equation}
    {Kurt}_{k} = \frac{\frac{1}{H\times W} \sum_{i=1}^{H} \sum_{j=1}^{W}  (\left | {\tilde{X} }_{k}^{:,:,i,j} -Mean_{k} \right |)^4}{{{^{Std_{k}4} } }  } 
\end{equation}

This aggregation process transforms localized discrepancy intensities into global statistical descriptors, thereby capturing holistic distribution patterns of forgery traces rather than relying on position-specific artifacts. Subsequently, the four types of differential statistical features are flattened and concatenated to form a global differential frequency representation $D\in \mathbb{R} ^{2304} $ as shown in equation 8, which effectively encodes vertical, horizontal, and intra-spectral anomaly patterns.

\begin{equation}
    \begin{split}
D= cat(flat({Mean}_{k} ),flat({Std}_{k}  ) ,flat ( {Skew}_{k} ) ,flat ( {Kurt}_{k} ) ) 
    \end{split}
\end{equation}

\subsubsection{Local Branch} 
\begin{figure}[htbp]
    \centering 
    \vspace*{10pt}
    \includegraphics[width=0.5\textwidth]{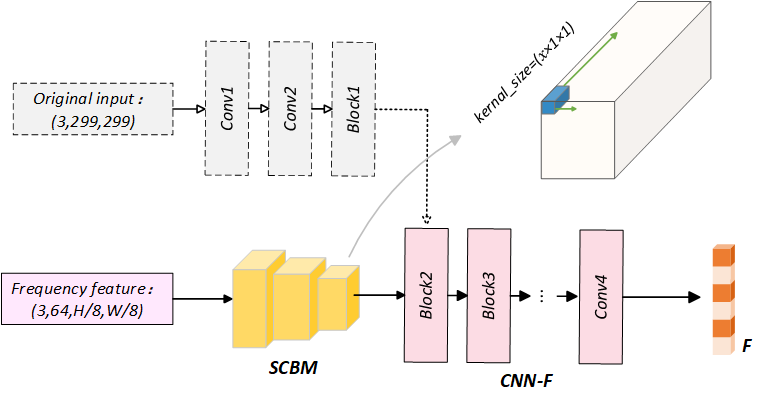} 
    \caption{The pipeline of Local Branch (inter/intra-block multi-scale frequency-convolutional network)} 
    \label{fig:Local Branch} 
\end{figure}

 2D CNNs exhibit inherent limitations in processing high-dimensional spectral features, failing to capture cross-frequency correlations (along the channel dimension) within individual image blocks. To address these limitations, we designed \textbf{Intra/Inter-block Multi-Scale Frequency-convolutional Network}, which consists of SBCM and CNN-F.The pipeline of Local Branch is depicted in Figure \ref{fig:Local Branch}.

2D CNNs exhibit inherent limitations in processing high-dimensional spectral features, as they fail to capture cross-frequency correlations (i.e., along the channel dimension) within individual image blocks. To overcome these challenges, we propose an Multi-Scale Frequency-Convolutional Network, which consists of a Spectral Band Convolution Module (\textbf{SBCM}) and a Frequency Convolutional Neural Network (\textbf{CNN-F}).

\textit{SBCM}. This module employs stacked 3D convolutional blocks with \textit{kernal\ size=(x,1,1)}  to capture inter-channel correlations among spectral coefficients. By sliding 3D kernels along the spectral band dimension, SBCM progressively learns localized interaction patterns across adjacent frequency bands. Initial layers utilize larger kernels to establish long-range dependencies among low-, mid-, and high-frequency components for cross-band feature aggregation;  intermediate layers progressively reduce \textit{x} to focus on abrupt spectral transitions caused by forgery artifacts, enhancing anomaly responses in mid-frequency regions; final layers adopt minimal \textit{x} to isolate high-frequency noise patterns and magnify distributional irregularities within critical spectral bands.   

The processed spectral-semantic features, denoted as $X_{f}^{local} \in \mathbb{R} ^{C\times 3\times (H/8)\times (W/8)}$ ,where C=64, thereby encapsulate hierarchically refined frequency-specific representation.

We then flatten the first two dimensions of the tensor, differing from the initial direct flattening approach, where each spatial location's 192-dimensional features now encapsulate high-level semantic characteristics with cross-band correlations, as shown in equation 9.
\begin{equation}
   X_{f}^{local} =flat(X_{f}^{local},dim=(0,1))
\end{equation}

\textit{CNN-F} To model the relationships between heterogeneous image blocks, we adapt conventional CNN architectures—exemplified by Xception—through modifications to the input layers. Specifically, we remove the first two convolutional layers and one depthwise separable convolutional block, replacing them with SBCM. To maintain architectural compatibility, we then adjust the input channels of the subsequent depthwise separable convolution block to match the channel dimension of the output spectral features $X_{f}^{local}$.
This hierarchical processing framework enables the modeling of both intra- and inter-block correlations among DCT coefficients, resulting in frequency-aware features $ F\in \mathbb{R}^{2048}$.

\subsection{Hierarchical Cross-Modal Fusion}
\subsubsection{Frequency-Aware Attention Enhancement}
\begin{figure}[htbp]
    \centering
    \includegraphics[width=\linewidth]{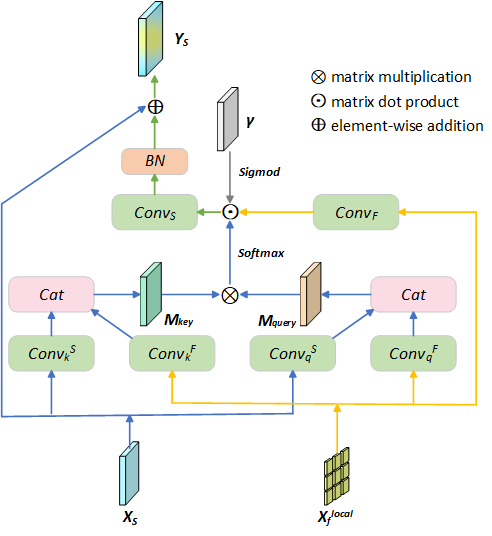}
    \caption{The structure of Frequency-Aware Attention Enhancement module.}
    \label{fig:faae}
\end{figure}

We aim to amplify localized forgery artifacts at the shallow layers of the spatial domain model, thereby enhancing the saliency of subtle manipulation traces to foster the learning of discriminative feature representations.The structure of Frequency-Aware Attention Enhancement module is shown in Figure \ref{fig:faae}.

The module initially receives two inputs: the shallow-layer spatial features $X_{S}$ and the spectral features of SBCM's output $X_{f}^{local} $. We first generate cross-modal query-key pairs through parallel convolutional projections, while aligning channel dimensions to ensure compatibility, as shown in equation 10 and 11.
\begin{equation}
    M_{query}=cat(Conv_{q}^{F}(X_{f}^{local} ),Conv_{q}^{S}(X_{S} ) ) 
\end{equation}
\begin{equation}
    M_{key}=cat(Conv_{k}^{F}(X_{f}^{local})^{T}  ,Conv_{k}^{S}(X_{S} )^{T} ) 
\end{equation}
The cross-modal attention weight matrix $\alpha$ is subsequently derived through scaled dot-product computation between the concatenated key pairs, which encapsulates the spectral-spatial interaction dynamics by correlating frequency-localized patterns with spatial anomalies, as shown in equation 12.
\begin{equation}
    \alpha =Softmax(\frac{M_{query}M_{key}^{T}  }{\sqrt{2H} } )
\end{equation}
 The spatial refinement branch injects frequency context constraints through attention gating. Here, $Conv_{F}$ projects spectral-domain features to align with the spatial feature dimensions, $\gamma _{S}$ denotes learnable parameters that dynamically regulate the contribution intensity of spectral features to the spatial branch,$\sigma $ represents the Sigmoid activation function, and $ Conv_{S} $ adapts the dimensionality of the frequency-aware attention matrix to ensure dimensional compatibility during cross-modal feature fusion, as shown in equation 13.
 \begin{equation}
     Y_{S}  =X_{S}+BN( Conv_{S}(\alpha \odot Conv_{F}( X_{f}^{local} )\cdot\sigma (\gamma _{S} ) ) )
 \end{equation}

\subsubsection{Hybrid Cross-Modal Attention Fusion} 
\begin{figure}[htbp]
    \centering
    \includegraphics[width=0.8\linewidth]{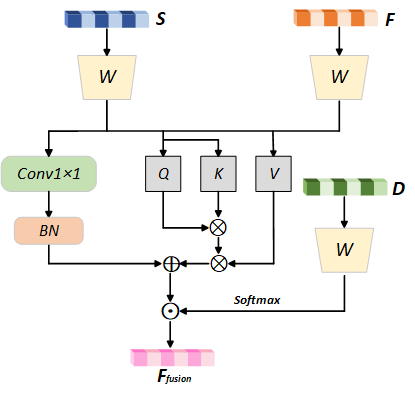}
    \caption{The structure of Hybrid Cross-Modal Attention Fusion.}
    \label{fig:finalfusion}
\end{figure}

To establish deep interactive reasoning between spatial and spectral domains, we propose an HCMA multi-modal fusion module, as is shown in Figure \ref{fig:finalfusion}. The module first projects spatial deep features $S\in \mathbb{R}^{1792}$ and intra/inter-block frequency correlation features $F\in \mathbb{R}^{2048}$ into a unified embedding space through linear transformation layers $S^{'},F^{'} \in \mathbb{R}^{1792}$, eliminating modality-specific dimensional discrepancies.

Using spatial features as Query vectors $Q=S^{'}W_{Q}$, and frequency features as Key-Value pairs $K=F^{'}W_{K}$,$V=F^{'}W_{V}$,where $W_{Q} , W_{K} ,W_{V} \in \mathbb{R} ^{1024\times 1024} $ are learnable parameter matrices. Then, multi-head attention is used to capture cross-modal global dependencies to generate attention-weighted representations, as shown in equation 14.
\begin{equation}
    A=softmax(\frac{QK^{T} }{\sqrt{1024/h} } )V
\end{equation}

To enhance fusion performance, we introduce residual connections by adding the attention output with 1D-convolution-processed spatial features followed by batch normalization, thereby maintaining gradient stability, as shown in equation 15.
\begin{equation}
    A_{res}=A+  BN(Conv_{1\times 1}(S^{'} ) ) 
\end{equation}

Considering the sensitivity of differential statistical features to manipulated regions, a dynamic gating mechanism is designed—\\projecting differential frequency statistical representation $D\in \mathbb{R}^{2304}$ into gating coefficients via Sigmoid activation, which modulates features to amplify anomaly responses, as shown in equation 16 and 17.
\begin{equation}
    g=\alpha (W_{g}D+b_{g}),W_{g}\in \mathbb{R}^{1024\times 2304}
\end{equation}
\begin{equation}
    F_{fused}=A_{res}\odot  g
\end{equation}

This architecture achieves hierarchical cross-modal feature synergy, combining global dependency modeling with local anomaly enhancement capabilities. Finally, the fused multi-modal semantic representation vectors are fed into the classifier to execute discriminative decision-making for deepfake detection tasks.

\section{Experiments}

\subsection{Implements Details}
For each video clip, we first extract the same amount of frames. Then, we employ the state-of-the-art face detector dlib to record facial bounding box coordinates for each image. During preprocessing, facial regions are cropped based on these coordinates (Face Tracker), and block-wise DCT transform with a block size of 8×8 is applied. The spatial pipeline and local stream of the frequency domain pipeline accepts inputs of 380×380 resolution, while the global stream of the frequency domain pipeline processes data into identical dimensions using the SIDA operations. In the model architecture, we adopt EfficientNet-B4 pre-trained on ImageNet as the backbone for the spatial-domain pipeline. In SCBM, 3D convolutional kernel sizes are set to 7, 5, and 3 sequentially. The HCMA module utilizes multi-head attention with 8 heads. For training, we optimize the model using the Adam optimizer with a learning rate of 0.001, weight decay of 1e-8, and a batch size of 20. The model is trained for 20 epochs on two NVIDIA RTX 4090 GPUs.

\begin{table}[htbp]
\caption{In-dataset evaluation results on the FF++ dataset with varying compression levels. The best results are highlighted in bold, and the second-best results are underlined.}
\begin{tabular}{l|cc|cc}
\hline
\multirow{2}{*}{Methods} & \multicolumn{2}{c|}{FF++(HQ)}                              & \multicolumn{2}{c}{FF++(LQ)}          \\
                         & \multicolumn{1}{l}{Acc(\%)} & \multicolumn{1}{l|}{AUC(\%)} & Acc(\%) & \multicolumn{1}{l}{AUC(\%)} \\ \hline
Xception\cite{xception}                 & 95.73                       & 96.30                        & 86.86   & 89.30                       \\
X-Ray\cite{x-ray}                    & -                           & 87.40                        & -       & 61.60                       \\
Two Branch\cite{twobranch}               & -                           & 98.70                        & -       & 86.56                       \\
SPSL\cite{spsl}                     & 91.50                       & 95.32                        & 81.57   & 82.82                       \\
MAD*\cite{mad}                      & \underline{97.17}                       & \underline{99.28}                             &87.35         &89.92                             \\
EDD\cite{edd}                       & 96.19                       & 98.81                        & 88.69       & \underline{91.27}                           \\ 
M2TR*\cite{m2tr}                     & 96.73                       &99.16                              &\underline{89.36}         &91.08                             \\
HFI-Net\cite{hfi-net}                  & 95.12                       & 98.66                        & 86.90   & 90.75                       \\
ED\cite{ed}                       & 94.98                       & 98.30                        & -       & -                           \\
\hline
SPCL-HCMF(ours)                     & \textbf{97.43 }                      & \textbf{99.58}                        &\textbf{91.77 }        &\textbf{94.21}                             \\ \hline
\end{tabular}
\label{tab:indataset}
\vspace{-0.5cm}
\end{table}

\subsection{Datasets}
Our experiments are primarily conducted on the FaceForensics++ (FF++) \cite{ff++} dataset.  FF++ comprises 1,000 original videos and 4,000 corresponding manipulated videos generated by four distinct forgery methods: Deepfakes, Face2Face, FaceSwap, and NeuralTextures.  The dataset includes three compression levels: c0 (raw/uncompressed), c23 (light compression), and c40 (heavy compression).  We focus on the c23 and c40 versions for evaluation.

\begin{table*}[htbp]
\caption{Performance comparison of different methods on various datasets. The best results are highlighted in bold, and the second-best results are underlined.}
\begin{tabular}{c|c|c|c|c|c}
\hline
Methods                                                                & Years & \begin{tabular}[c]{@{}c@{}}Training \\ dataset\end{tabular}  & \begin{tabular}[c]{@{}c@{}}FF++(c23)\\ AUC(\%)\end{tabular} & \begin{tabular}[c]{@{}c@{}}Celeb-DF(v2)\\ AUC(\%)\end{tabular} & \begin{tabular}[c]{@{}c@{}}DFDC\\ AUC(\%)\end{tabular} \\ \hline
Xception\cite{xception}        & 2017          &         & 96.30      & 65.30        &72.20       \\
Two-branch\cite{twobranch}     & 2020          &        & 93.18      & 73.41        &-       \\
X-Ray\cite{x-ray}           &2020           &       & 87.40     & -            & 70.00 \\
F\textsuperscript{3}-Net\cite{f3net}          & 2020          &FF++(c23)          & 98.10     & 65.17        &-       \\
MAD*\cite{mad}             & 2021          &       & 99.28       & 67.44        & 65.56     \\
M2TR*\cite{m2tr}            & 2022          &         & 99.16        & 65.17        & -     \\
SFDG\cite{dynamicgraph}            & 2023         &         & \underline{99.53}      & \textbf{75.83 }       & \underline{73.64} \\ 
DAW-FDD\cite{daw-fdd}              &2024           &         & 98.28     & 74.42        & 61.47      \\ 
\hline
SPCL-HCMF(ours)            & -         & FF++(c23)        & \textbf{99.58  }   & \underline{74.68}        & \textbf{73.71} \\ \hline
\end{tabular}%
\label{tab:crossdataset}
\vspace{-0.1cm}
\end{table*}

The Celeb-DF \cite{celebdf} dataset contains two versions (v1 and v2), featuring interviews with 59 celebrities across diverse ages, ethnicities, and genders.  The v1 subset includes 409 real videos and 795 synthesized videos, while v2 expands this to 590 real and 5,639 manipulated videos. We conducted cross-dataset evaluation using the enhanced v2 version, which incorporates a broader diversity of manipulation types.

The DFDC (DeepFake Detection Challenge) \cite{dfdc} dataset is a benchmark dataset comprising over 100,000 video clips sourced from 3,426 actors. In our experiments, the DFDC dataset is primarily used to evaluate the cross-dataset generalization capability of our model.

\subsection{Evaluation Metrics}
We adopt Accuracy (Acc) and Area Under the ROC Curve (AUC) as evaluation metrics, which are standard measures in deepfake detection research. All experimental results are reported at the image level.

\subsection{In-dataset evaluation}
To evaluate the effectiveness of the proposed method, we first perform intra-dataset experiments on the FaceForensics++ (\textbf{FF++}) dataset. Following the standard protocol, the dataset comprising 1,000 videos is partitioned into 720 for training, 140 for validation, and 140 for testing. For each video, 270 frames are extracted, and a four-fold augmentation is applied to balance the class distribution.

We conduct experiments on both the high-quality (HQ, c23) and low-quality (LQ, c40) compression settings, and report Accuracy (Acc) and Area Under the Curve (AUC) metrics in Table \ref{tab:indataset}. Our approach is benchmarked against state-of-the-art detection methods, including X-Ray \cite{x-ray}, MAD \cite{mad}, and M2TR \cite{m2tr}, etc. As shown in the results, our method achieves superior performance under both compression levels. In particular, when compared to frequency-aware baselines such as SPSL \cite{spsl} and HFI-Net \cite{hfi-net}, our approach improves detection accuracy by 5.9\% and 2.3\% on HQ, respectively. 

\subsection{Cross-dataset evaluation}
To demonstrate the generalization capability of our method, we conduct cross-dataset evaluations on multiple benchmark datasets. This setting presents greater challenges due to significant distribution shifts among datasets. Specifically, we train our model on FF++ (HQ) and evaluate it on Celeb-DF (v2) and DFDC. For each video, we extract 50 frames and compute the frame-level AUC scores. 
As shown in Table \ref{tab:crossdataset}, our method achieves superior or comparable performance across all cross-dataset evaluations. Notably, frequency-aware methods like F\textsuperscript{3}-Net \cite{f3net} report limitations in cross-datasets adaptability. In contrast, our approach outperforms them by 9.5\% AUC on Celeb-DF (v2), respectively. Compared to spatial-frequency fusion methods Two-branch \cite{twobranch} and SFDG \cite{dynamicgraph}, our method demonstrates improved robustness, justifying the effectiveness of the proposed hierarchical cross-modal fusion strategy in leveraging the complementary strengths of spatial and frequency features.

\subsection{Ablation Study}
In this section, we conduct ablation studies on our proposed key components — SCBM,  SIDA, and HCMF — to systematically investigate their contributions to detection performance. Specifically, we sequentially remove each component and evaluate the corresponding AUC scores on FF++ (c23), with results detailed in Table \ref{tab:abt1}.

\begin{table}[htbp]
\caption{Results of ablation experiments on three key components: SCBM, HCMF and SIDA.}
\begin{tabular}{ccc|c}
\hline
SCBM                      & HCMF                      & SIDA                        & AUC(\%)   \\ \hline
-                         & -                         & -                         & 97.70 \\
-                         & \checkmark & \checkmark & 99.03 \\
\checkmark & -                         & \checkmark &98.95       \\
\checkmark & \checkmark & -                         & 97.99      \\
\checkmark & \checkmark & \checkmark & 99.58 \\ \hline
\end{tabular}
\label{tab:abt1}
\vspace{-0.1cm}
\end{table}

\textit{Ablation Study on SCBM.} SCBM is designed to capture subtle frequency-based forgery traces within individual image blocks. To evaluate its effectiveness, we removed SCBM from the local branch in the frequency domain while keeping all other modules unchanged, directly feeding the frequency-domain features into CNN-F.  As shown in Table \ref{tab:abt1}, this modification resulted in a 0.5\% drop in performance, confirming SCBM’s ability to capture abrupt variations in DCT coefficients and enhance local spectral feature representation.

\textit{Ablation Study on Hierarchical Cross-Modal Fusion}. The proposed hierarchical fusion framework consists of FAAE module and HCMA module, which jointly model and integrate spatial and multi-scale frequency-domain features. Replacing this mechanism with a simple concatenation of spatial and frequency features led to a 0.6\% performance decrease, demonstrating the effectiveness of our design. Further ablation was conducted by individually removing the FAAE and HCMA modules. As shown in Table \ref{tab:abt2}, the framework achieves optimal performance only when both modules are jointly employed, highlighting their complementary roles in enhancing cross-modal feature fusion.

\begin{figure}[htbp]
    \centering
    \includegraphics[width=\linewidth]{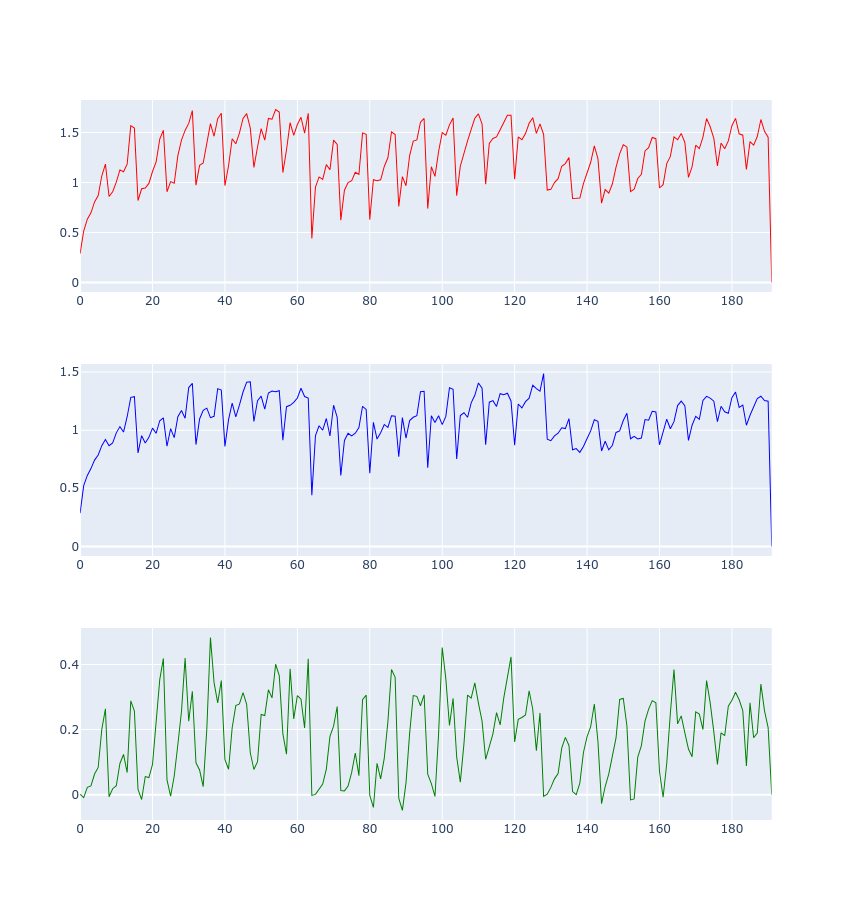}
    \caption{The line charts of the differential statistical metrics in SIDA.  Red represents authentic faces, blue denotes forged faces, and green indicates the difference between them.}
    \label{fig:diff_plot}
\end{figure}

\textit{Ablation Study on SIDA.} The SIDA aims to fully leverage global frequency differential statistical characteristics. By removing the SIDA (Glocal Branch) in the frequency domain and retaining only the hierarchical cross-modal fusion of spatial features and local frequency-domain features. As shown in Table \ref{tab:abt1}, performance declined by 1.4\%. This validates the effectiveness of the SIDA in feature refinement.

To provide a more intuitive visualization of SIDA's effects, we plotted line charts (using the mean as an example and considering only the Y component) for the differential statistical metrics in SIDA.   As shown in Figure \ref{fig:diff_plot},the 192-dimensional feature space consists of three segments: the first 64 dimensions correspond to inter-block row differences, the middle 64 dimensions represent inter-block column differences, and the last 64 dimensions indicate intra-block differences.  The results demonstrate that SIDA effectively amplifies the distinctions between authentic and forged images in mid-to-high frequency ranges.
\begin{table}[htbp]
\caption{Results of ablation experiments on key components (FAAE and HCMA) of HCMF}
\begin{tabular}{cc|c}
\hline
FAAE & HCMA & AUC(\%)   \\ \hline
-     & -   & 97.70 \\
\checkmark   & -   & 98.30 \\
-     &\checkmark    & 99.27      \\
\checkmark     & \checkmark   &99.58      \\ \hline
\end{tabular}
\label{tab:abt2}
\vspace{-0.5cm}
\end{table}

\section{Conclusion}
In this paper, we introduced a Spatial-Frequency Collaborative Learning framework with Hierarchical Cross-Modal Fusion for robust deepfake detection.
First, a block-wise DCT analysis with inter/intra-block multi-scale frequency-convolutional network was proposed to enhance local representation learning and enable precise forgery localization in the frequency domain.
Second, a scale-invariant differential analysis approach was designed to capture holistic forgery traces while mitigating multi-scale discrepancies and post-processing interference.
Finally, a hierarchical cross-modal fusion mechanism unified shallow-layer attention refinement and deep-layer dynamic modulation, addressing spatial-frequency  interdependencies and feature heterogeneity across modalities.
Extensive in-dataset and cross-dataset experiments demonstrate the superior detection performance and generalization capability of our method.  Ablation studies further validate the critical role of the  various key components in our design.

\begin{acks}
To Robert, for the bagels and explaining CMYK and color spaces.
\end{acks}

\bibliographystyle{ACM-Reference-Format}
\bibliography{sample-base}

\end{document}